\pdfoutput=1

\documentclass[sigconf]{acmart}
\usepackage{soul}
\usepackage{threeparttable}
\usepackage{multirow}

\copyrightyear{2024}
\acmYear{2024}
\setcopyright{acmlicensed} 
\acmDOI{xxxxxx}

\acmConference[SIGIR-AP '24]{Proceedings of the 2024 Annual International ACM SIGIR Conference on Research and Development in Information Retrieval in the Asia Pacific Region}{December 9--12, 2024}{Tokyo, Japan}

\acmBooktitle{Proceedings of the 2024 Annual International ACM SIGIR Conference on Research and Development in Information Retrieval in the Asia Pacific Region (SIGIR-AP '24), December 9--12, 2024, Tokyo, Japan}
\acmISBN{979-8-4007-0724-7/24/12}

\begin{document}

\title{Optimizing LLMs with Direct Preferences: A Data Efficiency Perspective}

\author{Pietro Bernardelle}
\orcid{0009-0003-3657-9229} 
\affiliation{%
  \institution{The University of Queensland}
  \city{Brisbane}
  \country{Australia}
}
\email{p.bernardelle@uq.edu.au}

\author{Gianluca Demartini}
\orcid{0000-0002-7311-3693} 
\affiliation{%
  \institution{The University of Queensland}
  \city{Brisbane}
  \country{Australia}
}
\email{demartini@acm.org}

\renewcommand{\shortauthors}{Pietro Bernardelle and Gianluca Demartini}

\begin{abstract}
Aligning the output of Large Language Models (LLMs) with human preferences (e.g., by means of reinforcement learning with human feedback, or RLHF) is essential for ensuring their effectiveness in real-world scenarios.
Despite significant advancements in LLM alignment techniques, the impact of different type of preference data on model performance has yet to be systematically explored.
In this study, we investigate the scalability, data efficiency, and effectiveness of Direct Preference Optimization (DPO) in fine-tuning pre-trained LLMs, aiming to reduce their dependency on extensive amounts of preference data, which is expensive to collect. We (1) systematically compare the performance of models fine-tuned with varying percentages of a combined preference judgement dataset to define the improvement curve of DPO and assess its effectiveness in data-constrained environments; and (2) provide insights for the development of an optimal approach for selective preference data usage.
Our study reveals that increasing the amount of data used for training generally enhances and stabilizes model performance. Moreover, the use of a combination of diverse datasets significantly improves model effectiveness. Furthermore, when models are trained separately using different types of prompts, models trained with conversational prompts outperformed those trained with question answering prompts.
\end{abstract}



\maketitle

\section{INTRODUCTION}
The rise of LLMs, such as OpenAI’s GPT \cite{GPT} and Google’s BERT \cite{BERT} families, has marked a revolutionary advancement in natural language processing.
These models excel in syntactic and semantic understanding yet aligning them with human preferences remains challenging. 

Over the past few years, significant work has attempted to address the misalignment challenge. This has led to the adoption of Reinforcement Learning (RL) techniques that incorporate human preferences to guide LLMs optimization. 
\citet{DPO} introduced one of the most innovative approach in this field: Direct Preference Optimization (DPO). Despite its promise, to date, only a limited
number of studies have explored its applications and potential benefits, suggesting that much remains to be discovered and understood about its effectiveness. This gap in research highlights the need to further investigate how DPO can enhance the alignment of language models with human preferences efficiently and effectively.

Our research conducts an experimental exploration of the DPO
method, examining its nuances and evaluating its efficacy in real-world applications. The objective is to improve our understanding of the impact of DPO, setting the stage for future development of efficient methods for utilizing human preference data and streamlining the LLM training process. To achieve our goal, we aim to address the following research questions:

\textbf{RQ1}:  How does the performance of LLMs fine-tuned with DPO evolve as increasingly larger subsets of preference data are used for training?

\textbf{RQ2}: How does the nature of the training data, specifically conversational versus question answering datasets, impact the model performance under DPO?

\begin{table*}[ht!]
\centering
\begin{threeparttable}
\caption{Overview of the datasets used in the analysis. The table details dataset size, partitioning into training, evaluation, and testing sets, and the types of prompts included.}
\begin{tabular}{c|ccccc}
\toprule
\textbf{Dataset} & \textbf{Size} & \textbf{Training (80\%)} & \textbf{Evaluation (10\%)} & \textbf{Testing (10\%)} & \textbf{Prompt Type} \\
\midrule
Dataset A\tnote{\footnotesize{a}} & 7,560 & 6,048 & 756 & 756 & Conversational \\
Dataset B\tnote{\footnotesize{b}} & 12,900 & 10,320 & 1,290 & 1,290 & Question-Answering \\
Dataset C\tnote{\footnotesize{c}} & 63,600 & 50,880 & 6,360 & 6,360 & Question-Answering \\
\midrule
\multirow{2}{*}{Combination} & \multirow{2}{*}{84,060} & \multirow{2}{*}{67,248} & \multirow{2}{*}{8,406} & \multirow{2}{*}{8,406} & Conversational \& \\
& & & & & Question-Answering \\
\bottomrule
\end{tabular}
\begin{tablenotes}
\footnotesize{
\item[a] \url{https://huggingface.co/datasets/argilla/distilabel-capybara-dpo-7k-binarized}
\item[b] \url{https://huggingface.co/datasets/argilla/distilabel-intel-orca-dpo-pairs}
\item[c] \url{https://huggingface.co/datasets/argilla/ultrafeedback-binarized-preferences}}
\end{tablenotes}
\label{tab:datasets_characteristics}
\end{threeparttable}
\end{table*}

\section{RELATED WORK}
In recent years, numerous influential studies have tackled the challenge of misalignment in language models. \citet{RLHF} are the pioneers in
merging the concepts of Reinforcement Learning
(RL) with Human Feedback (RLHF), showcasing the
potential of this method in guiding the training of
models with direct human preference judgements. Their work
laid the foundation for future enhancements in model
behavior through human evaluation.
Building on these foundational insights, \citet{ouyang2022training} and  \citet{bai2022training} extended the application of
RLHF to a broader range of language tasks, further
developing the methodology for LLMs. Their
contributions have been crucial in detailing the
operational framework of RLHF, which involves a
three-phase alignment process: (1) pretraining and supervised fine-tuning of the language model; (2) training of a reward model using explicit human preferences; and (3) alignment of the language model via
reinforcement learning through the reward model.

While effective in some scenarios, questions arise
about the efficiency, scalability, and extensive data
requirements of this approach during the training process.
Recent developments in the field have introduced a
significant paradigm shift, aiming to mitigate these
challenges.
\citet{DPO} introduced a 
methodology known as DPO, which seeks to optimize
language model alignment directly based on human
preferences, thus eliminating the need for an explicit
reward model (having it embedded in a loss function) and for reinforcement learning.
By eliminating the reliance on reinforcement
learning, this approach streamlines the training
process and challenges the conventional wisdom of
reinforcement learning-based alignment frameworks.
Although there have been considerable advancements in language model alignment techniques, focusing primarily on refining architectures for improved efficiency, the crucial aspect of human preference data utilization remains largely under-explored.

\section{METHODOLOGY}

\paragraph{\textbf{Data}}
To address the research questions outlined above, we conducted our experiments using three open-source preference judgement datasets. Table~\ref{tab:datasets_characteristics} provides an aggregated view of their characteristics.  
All selected datasets are sourced from Hugging Face and provided by Argilla\footnote{\url{https://huggingface.co/argilla}}. Two of the three datasets employed in our study originate from established datasets: OpenOrca \cite{OpenOrca} and UltraFeedback \cite{cui2023ultrafeedback}. The third dataset is based on a newly curated collection featuring conversational style prompts: Capybara\footnote{\url{https://huggingface.co/datasets/LDJnr/Capybara}}. The selected datasets provide a broad spectrum of scenarios and data characteristics, enabling a thorough assessment of the DPO-aligned models’ behavior when exposed to different volumes and types of preference data.

\paragraph{\textbf{Experimental setup}}
Our study is structured into two experiments.
The first experiment, addressing RQ1, is designed to determine the performance of
DPO-aligned models as the amount of data increases.
We combined the individual datasets into a single pool to control for content variability\footnote{The pool, formed by combining the entire individual datasets, was then split into training, evaluation, and testing segments, allocated as 80\%, 10\%, and 10\% respectively.}. From this combined dataset, we randomly sampled five subsets—20\%, 40\%, 60\%, 80\%, and 100\%—of the training split. Each subset was used to train separate instances of the base model. OpenHermes-2.5-Mistral-7B\footnote{\url{https://huggingface.co/teknium/OpenHermes-2.5-Mistral-7B}} was adopted as the base model, primarily to ensure our experiments remain reproducible and grounded in a high-performing, state-of-the-art open-source model. 
The process was repeated three times with different random seeds, and resulted in a total of 15 different DPO-aligned models. By incrementally increasing the data volume
and repeating the training multiple times, we accounted for variability and aimed to smooth the results by averaging them.

The second experiment, addressing RQ2, focuses on the individual characteristics of the three datasets. This approach is designed to discern whether specific types of data, particularly those with conversational versus question answering style prompts, have a more pronounced effect on the efficiency and effectiveness of the training under DPO. 
In the second experiment, we mirrored the methodology used in the initial experiment but applied it to each individual dataset separately. This approach yielded 45 distinct DPO-aligned models.

\paragraph{\textbf{Performance Evaluation}} Each DPO-aligned model was evaluated relative to the base model
using MT-Bench \cite{zheng2024judging}. MT-Bench presents a series of questions to both the base model and the DPO-aligned model, and for each question, it determines a `win' for the model that provides the best answer or records a `tie' if no clear best answer is identified. Based on the structured evaluation framework provided by MT-Bench, we calculated the improvement over the base model   by subtracting the
percentage of wins of the base model from the
percentage of wins of the model aligned with DPO,
as the amount of data used for alignment increases.
Additionally, we defined and calculated the tie rate between the DPO-aligned model and the base model at each data percentage subset as follows:
\[
\text{Tie Rate} = 1 - (\text{Win Rate} + \text{Lose Rate})
\]
where the win rate and lose rate are derived from the combined results of the three runs for each data percentage subset.

\paragraph{\textbf{Computational Resources}}
For our experiments, we utilized a single H100 GPU card and approximately 80GB
of RAM to accommodate the datasets. The training duration varied depending on
the dataset size. For models relying on the largest datasets, the training process
took up to one day, while models trained on the smallest datasets completed in
approximately two hours.

\section{RESULTS}

\subsection{Defining DPO’s Improvement Curve}

\begin{figure}
    \centering
\includegraphics[width=1\linewidth]{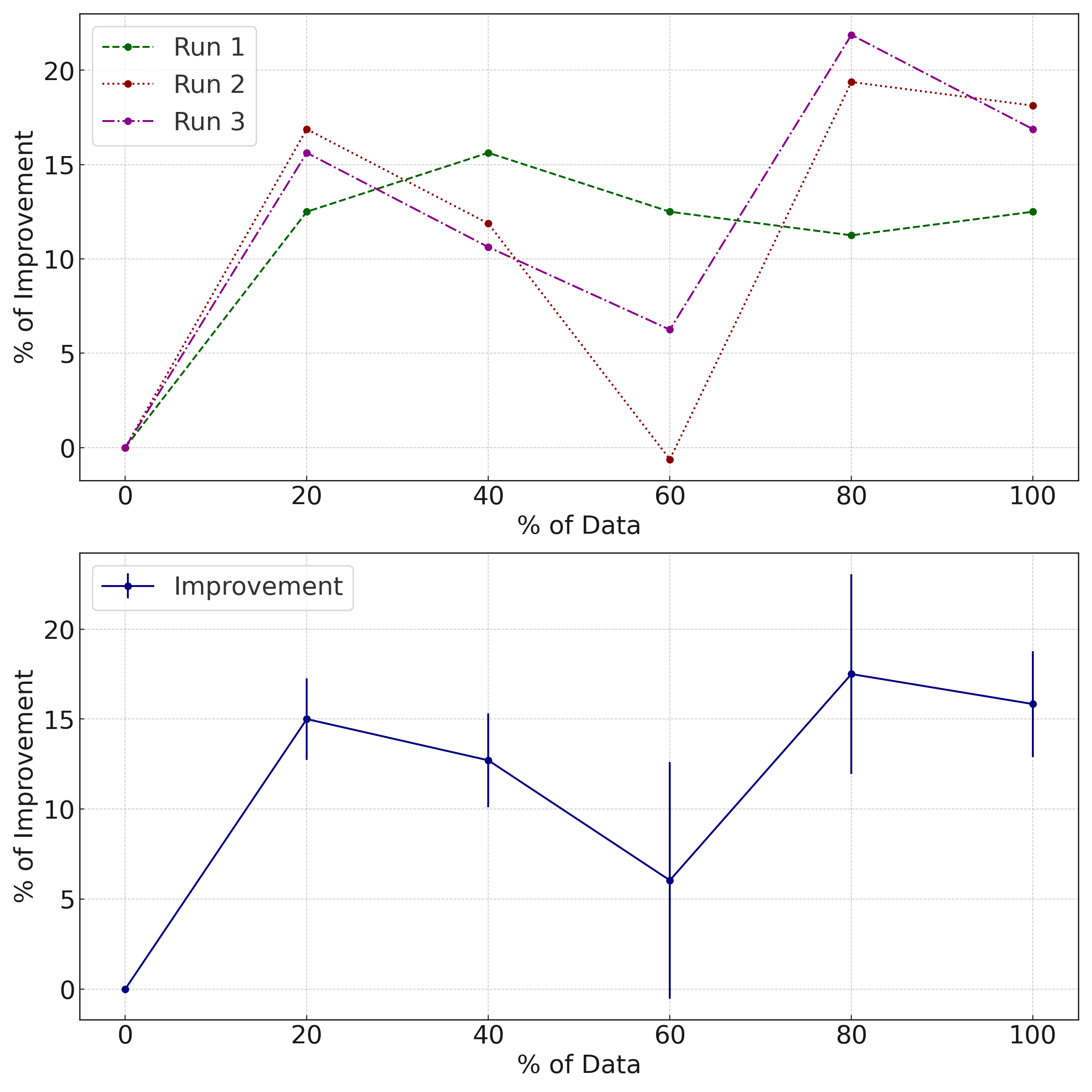}
    \caption{The top plot illustrates the improvement curves of DPO across three different experimental runs using the combined dataset. The bottom plot presents the averaged improvement curve of DPO, aggregating the results from the three experimental runs. Error bars indicate the standard deviation across the runs.}
    \label{fig:experiment_1_runs}
\end{figure}

Figure \ref{fig:experiment_1_runs} illustrates the percentage  improvement
in model performance across the three random seeds for the first experiment. 
Based on the results summarized in Figure \ref{fig:experiment_1_runs},
there is noticeable variability in performance
improvement across the three experimental runs,
indicating that the models' performance may vary
significantly depending on the specific data subsets
used for alignment. Runs 2 and 3 exhibit similar
patterns (Pearson $R = 0.94$, $P = .006$, $\alpha = .05$), diverging noticeably from run 1, especially
in the middle data usage percentages (40\%, 60\% and
80\%). 
The similarity between runs 2 and 3 compared to run 1, coupled
with the marked dip at 60\% followed by a significant
rebound, highlights the need for a more detailed
analysis into how different data samples may distinctly
affect the training process.
Which preference judgements we use does matter.
A comprehensive examination of
these data subsets could reveal critical insights into
optimizing the training process for more consistent
and predictable improvements.
Despite these fluctuations, a consistent trend
emerges in which increased data usage generally
correlates with enhanced performance improvements.
This pattern supports the
hypothesis that greater data volumes positively
influence the model's alignment with human
preferences. However, the overall trend is not as smooth as initially hypothesized, which may be attributed to the variability across the three runs. Additional runs could potentially smooth out these irregularities, providing a clearer depiction of the trend.

Additionally, it is interesting to point out that, as
the data volume increases, the model aligned with
DPO becomes distinctly more accurate in
its responses compared to the base model. 
The tie rate between models decreases as the percentage of data used for DPO alignment increases, suggesting that the judging mechanism finds it easier to discern and prefer the responses of the DPO-aligned model. This indicates clearer differentiation and better alignment with desired outputs.

\begin{figure}
    \centering
    \includegraphics[width=1\linewidth]{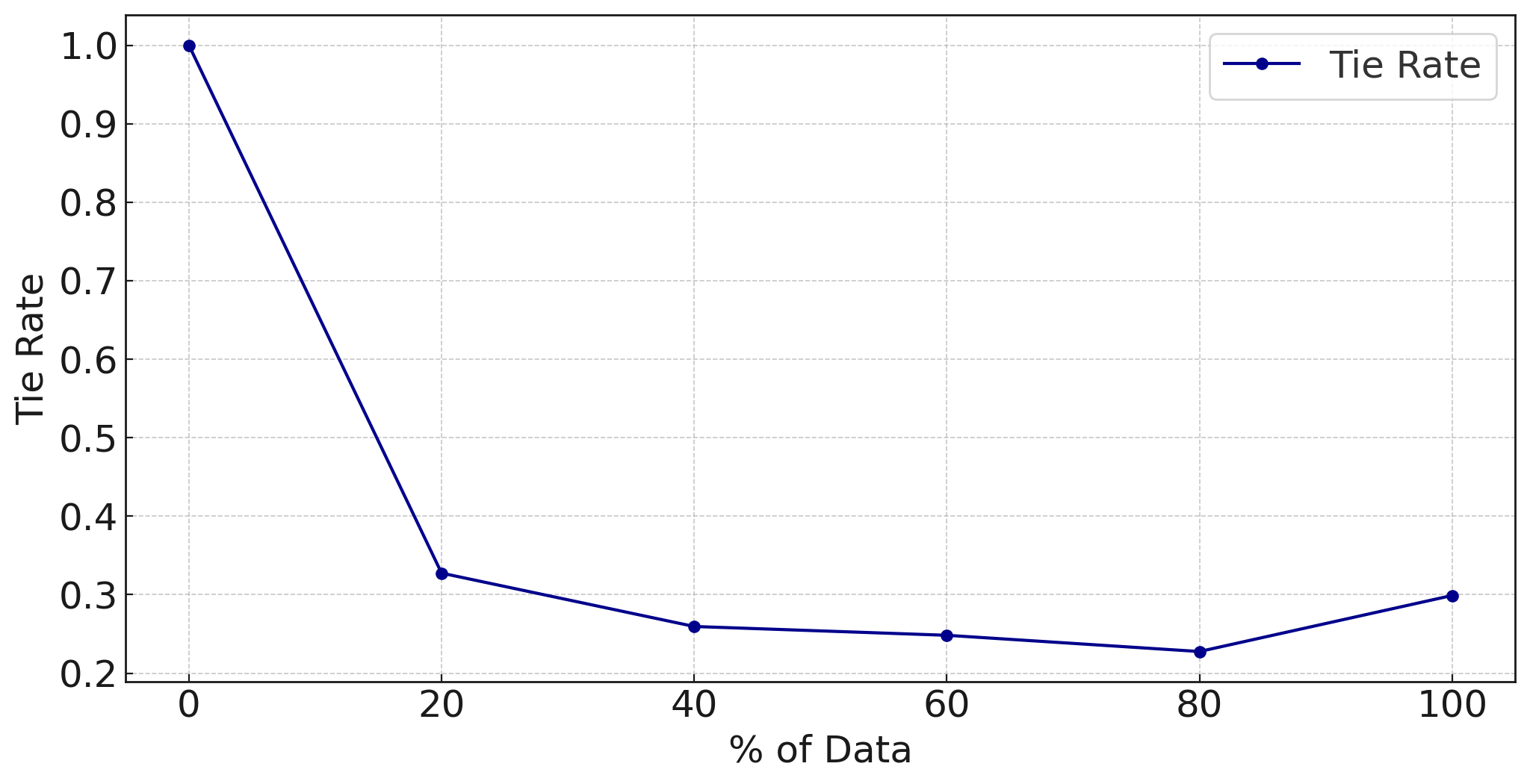}
    \caption{Relationship between the percentage of data and tie rate. The plot demonstrates how the tie rate tend to decrease as the percentage of data increases.}
    \label{fig:tie_rate}
\end{figure} Figure~\ref{fig:tie_rate} reveals that the tie rate consistently diminishes as the data volume increases, with the exception of the 100\% data usage mark. This pattern provides two significant insights:

\begin{enumerate}
    \item The judging model is convinced of the performance dip observed around 60\%
of data usage shown in Figure~\ref{fig:experiment_1_runs}, indicating that at this stage, the DPO-aligned model responses
are less distinguishable from the base model responses.
    \item The increased tie rate at 100\% data usage suggests that the judging model
confidence in distinguishing between the DPO-aligned and base model decreases
when the improvement curve plateaus. This phenomenon implies
that while additional data contributes to performance gains, there might be a
threshold beyond which the added data does not significantly enhance model
differentiation. 
\end{enumerate}

\subsection{Impact of Data Nature}

\begin{figure}
    \centering
    \includegraphics[width=1\linewidth]{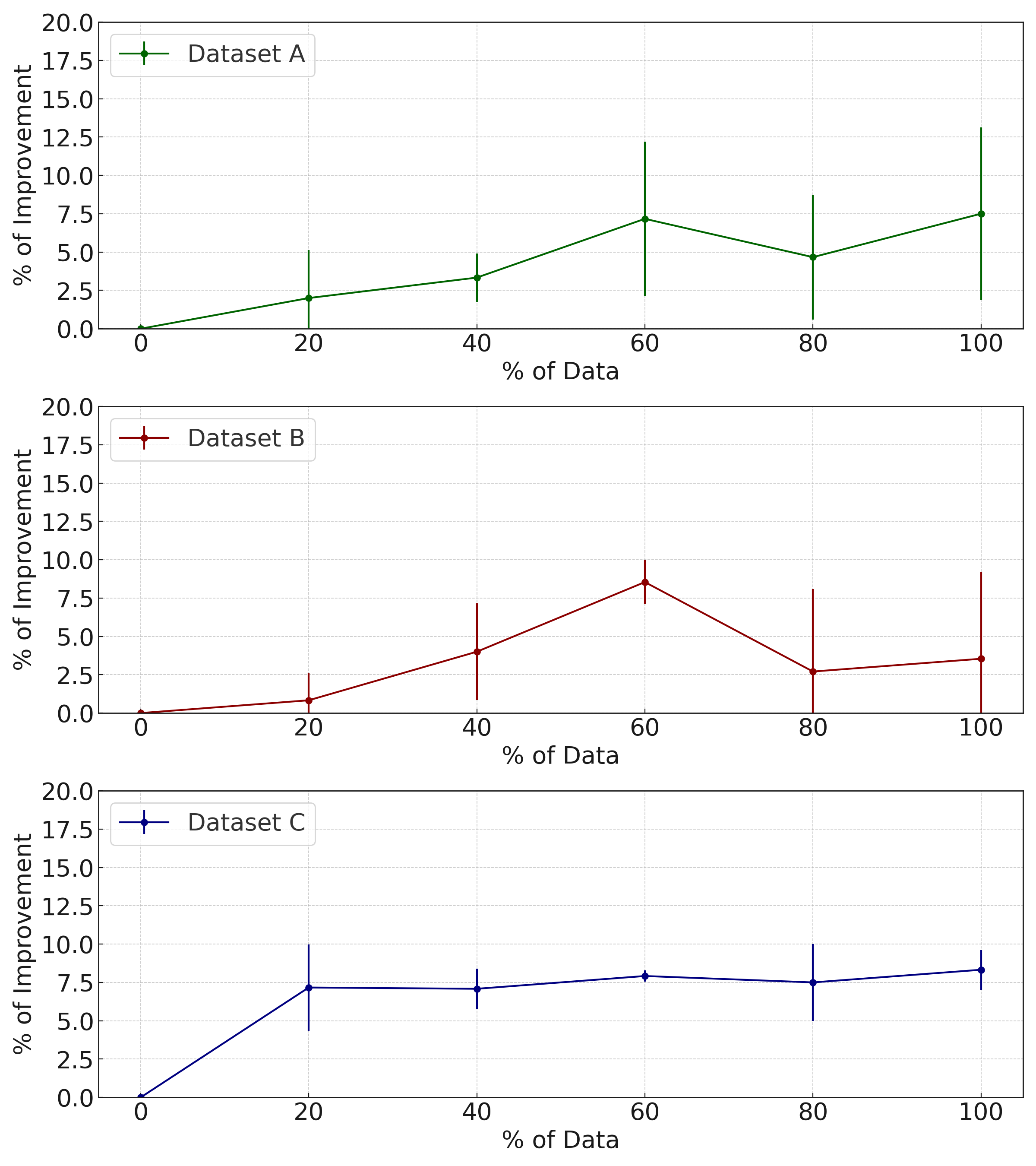}
    \caption{Comparison of the percentage of improvement
across three datasets (Dataset A, Dataset
B, Dataset C) relative to the percentage of data used for
alignment.}
    \label{fig:experiment_2_trend}
\end{figure}

When we delve into the unique
characteristics of the three individual datasets, we can identify the distinct contributions and impacts of different data
types on model performance. As Figure~\ref{fig:experiment_2_trend} shows:

\begin{itemize}
    \item  Dataset A, the smallest dataset in our study, shows a positive trend in model performance with increased data usage, achieving improvements comparable to those of models trained on much larger datasets. This highlights the significant role that conversational prompts play in providing dynamic and context-rich interactions.
    \item Dataset B shows the potential for significant performance
improvements through DPO, albeit with some variability and non-linear trends.
The unexpected dip in performance at intermediate data volumes underscores the
need for careful curation of training data to maximize the benefits of additional data. 
    \item Dataset C large size allows for more robust learning, reducing the impact
of any individual data points that might introduce noise or anomalies. The 
improvement curve suggests that the model effectively leverages the additional data,
steadily refining its alignment with human preferences as more data is introduced.
This contrasts with the more variable performance observed in the smaller datasets,
where the model improvements were less consistent and more sensitive to specific
subsets of data.
\end{itemize}

\subsection{Implications and Observations}

\begin{table}[t]
\centering
\caption{Performance improvements of the DPO-aligned models across the four
datasets. The table lists both peak and average percentage improvements compared to the base model without alignment.}
\begin{tabular}{c|cc}
\toprule
\multirow{2}{*}{\textbf{Dataset}} & \textbf{Peak} & \textbf{Avg}\\[-0.5ex]  & \textbf{Improvement (\%)} & \textbf{Improvement (\%)}\\
\midrule
Dataset A & 7.5\% & 4.93\%\\
Dataset B & 8.54\% & 3.925\%\\
Dataset C & 8.325\% & 7.6\%\\
Combination & 19.126\% & 15.35\%\\
\bottomrule
\end{tabular}
\label{tab:improvement}
\end{table}

The results from the two experiments provide several important implications for the application of DPO in language model training:
\begin{itemize}
    \item \textbf{Significance of Data Volume}: The  trend across all datasets
highlights that increased data volume for alignment generally enhances  model performance
and stability.
    \item \textbf{Combined Datasets for Alignment}: Combining multiple datasets for alignment consistently results in superior
performance compared to using individual datasets, as shown in Table~\ref{tab:improvement}. This finding underscores
the importance of diversity and comprehensiveness in training data, suggesting
that a holistic approach to dataset selection can significantly enhance model
effectiveness. 
    \item \textbf{Impact of Data Type}: The type of data significantly influences the model improvement. Conversational prompts (Dataset A) lead to steady enhancements
despite its smaller size, likely because the model can grasp a better
understanding of the context with this type of data. This indicates that conversational
prompts are particularly effective in improving model performance
and could be prioritized in future data selection approaches.
    \item \textbf{Performance Dips and Training Dynamics}: The observed dips in performance across different data volumes, though not entirely explainable through
our current analysis, point to the inherent complexities in training dynamics.
These fluctuations suggest that there are underlying factors affecting performance
that warrant further investigation. The random sampling of data likely
contributed to the dispersion of noise across the performance curves, indicating a need for more controlled and systematic data sampling methods in future
studies.
\end{itemize}

This finding underscores the value of using a more extensive and varied dataset
for alignment, as it can capture a broader range of linguistic patterns and nuances, thereby enhancing the overall performance of the model.

\section{CONCLUSION AND FUTURE WORK}
In this study, we have explored the effectiveness and efficiency of DPO in aligning LLMs to human preference judgments. Our findings indicate that the relationship
between data volume and model performance improvement is not strictly linear, as might be expected; rather, it shows notable fluctuations
influenced heavily by the specific data subsets being used. 
Thus, preference judgement quality over quantity might be the way forward for LLM alignment.
Overall, this research enriches our knowledge of the DPO method, highlighting the need for data selection strategies that can systematically and consistently yield favorable results, even with limited amounts of data.

There is considerable potential in developing targeted strategies for optimizing preference data selection, which could streamline LLM training processes, enhancing efficiency and reducing costs. 
By pursuing this direction, future research can continue to refine and enhance the alignment of LLMs with human preferences, ultimately advancing natural language processing technologies.

\begin{acks}
This work is partially supported by the Australian Research Council (ARC) Training Centre for Information Resilience (Grant No. IC200100022).
\end{acks}

\bibliographystyle{ACM-Reference-Format}
\bibliography{optimising-LLMs-dpo}

\end{document}